\documentclass[10pt,twocolumn,letterpaper]{article}

\usepackage[pagenumbers]{cvpr} 

\usepackage{graphicx}
\usepackage{amsmath}
\usepackage{amssymb}
\usepackage{booktabs}
\usepackage{multirow}

\usepackage[pagebackref,breaklinks,colorlinks]{hyperref}

\usepackage[capitalize]{cleveref}
\crefname{section}{Sec.}{Secs.}
\Crefname{section}{Section}{Sections}
\Crefname{table}{Table}{Tables}
\crefname{table}{Tab.}{Tabs.}

\def\cvprPaperID{1134} 
\def\confName{CVPR}
\def\confYear{2023}

\begin{document}

\title{Ambiguity-Resistant Semi-Supervised Learning for Dense Object Detection}

\author{
Chang Liu\textsuperscript{1,}\footnotemark[1] , 
Weiming Zhang\textsuperscript{2,}\footnotemark[1] , 
Xiangru Lin\textsuperscript{2},
Wei Zhang\textsuperscript{2,}\footnotemark[2] , 
Xiao Tan\textsuperscript{2}, 
Junyu Han\textsuperscript{2}, \\
Xiaomao Li\textsuperscript{1,3},
Errui Ding\textsuperscript{2},
Jingdong Wang\textsuperscript{2} \\
\textsuperscript{1} Shanghai University,
\textsuperscript{2} Baidu Inc,
\textsuperscript{3} Shanghai Artificial Intelligence Laboratory \\
\tt\small 
\{liuchang123,lixiaomao\}@shu.edu.cn \\
\tt\small 
\{zhangweiming,linxiangru,zhangwei99,tanxiao01,hanjunyu,dingerrui,wangjingdong\}@baidu.com
}

\maketitle
\renewcommand{\thefootnote}{\fnsymbol{footnote}}
\footnotetext[1]{Co-first author (Equal Contribution).} 
\footnotetext[2]{Corresponding author. \\
This work was done when Chang Liu was an intern at Baidu Inc.}

\begin{abstract}
With basic Semi-Supervised Object Detection (SSOD) techniques, one-stage detectors generally obtain limited promotions compared with two-stage clusters.
We experimentally find that the root lies in two kinds of ambiguities: 
(1) Selection ambiguity that selected pseudo labels are less accurate, since classification scores cannot properly represent the localization quality. 
(2) Assignment ambiguity that samples are matched with improper labels in pseudo-label assignment, as the strategy is misguided by missed objects and inaccurate pseudo boxes.
To tackle these problems, we propose a \textbf{Ambiguity-Resistant Semi-supervised Learning} (ARSL) for one-stage detectors.
Specifically, to alleviate the selection ambiguity, Joint-Confidence Estimation (JCE) is proposed to jointly quantifies the classification and localization quality of pseudo labels.
As for the assignment ambiguity, Task-Separation Assignment (TSA) is introduced to assign labels based on pixel-level predictions rather than unreliable pseudo boxes.
It employs a ’divide-and-conquer’ strategy and separately exploits positives for the classification and localization task, which is more robust to the assignment ambiguity.
Comprehensive experiments demonstrate that ARSL effectively mitigates the ambiguities and achieves state-of-the-art SSOD performance on MS COCO and PASCAL VOC.
Codes can be found at \href{https://github.com/PaddlePaddle/PaddleDetection}{https://github.com/PaddlePaddle/PaddleDetection}.
\end{abstract}

\section{Introduction}
\label{sec1}

\begin{figure}[t]
    \centering
    \setlength{\abovecaptionskip}{5pt}
    \setlength{\belowcaptionskip}{-10pt}
    \includegraphics[width=\linewidth]{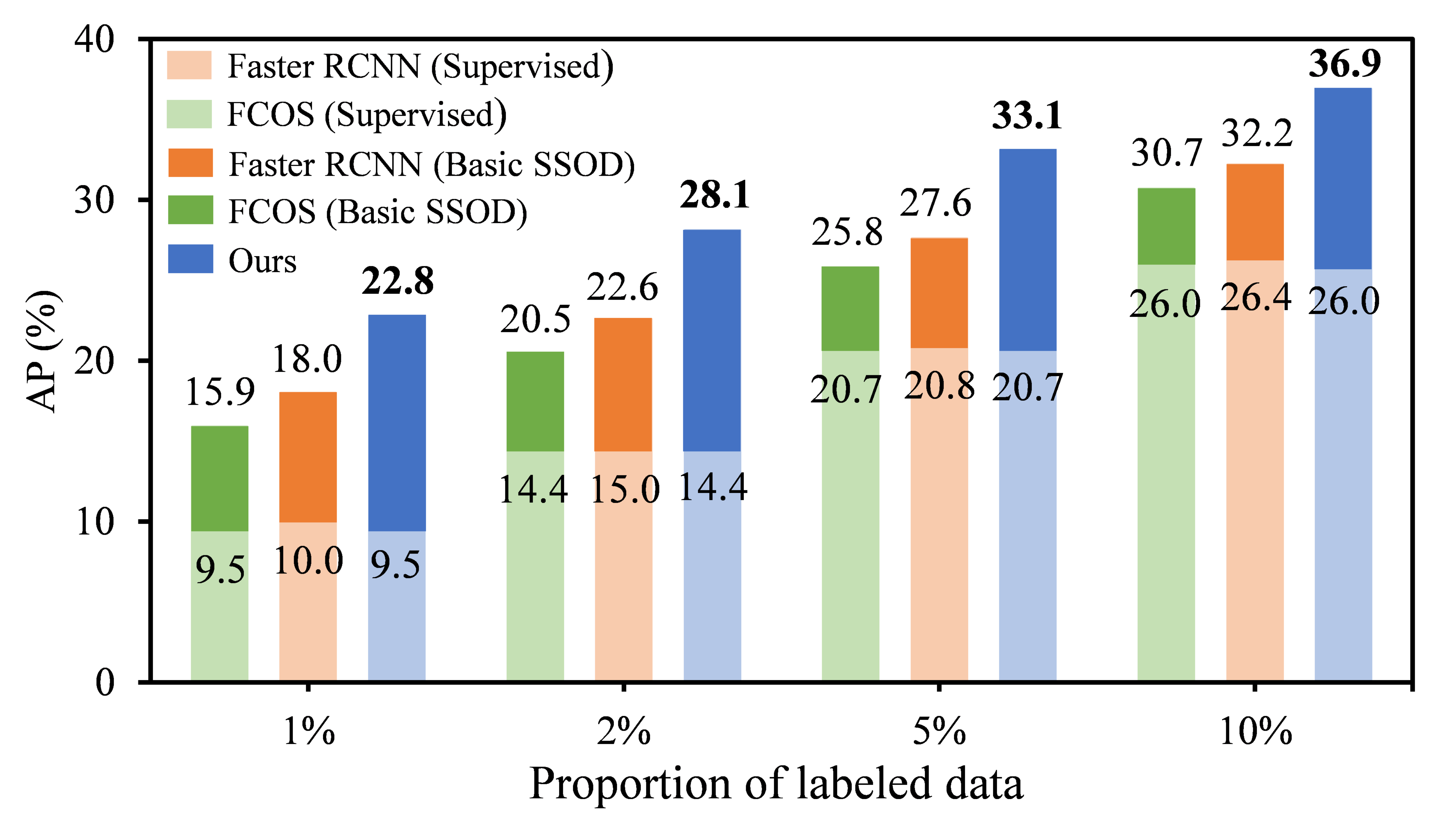}
    \caption{Comparing FCOS, Faster RCNN, and our approach on COCO \emph{train2017}. Under the basic SSOD pipeline, FCOS obtains limited improvements compared with Faster RCNN. Our approach consistently promotes FCOS and achieves a state-of-the-art performance on SSOD.}
    \label{fig1}
\end{figure}

Abundant data plays an essential role in deep learning based object detection \cite{FasterRCNN,Ssd,Yolo}, yet labeling a large amount of annotations is labour-consuming and expensive. 
To save labeling expenditure, Semi-Supervised Object Detection (SSOD) attempts to leverage limited labeled data and easily accessible unlabeled data for detection tasks.
Advanced SSOD methods \cite{UnbiasedTeacher,SoftTea} follow the Mean-Teacher \cite{MeanTea} paradigm and mainly apply the self-training \cite{PseudoLabel,NoisyStudent} technique to perform semi-supervised learning. 
Though this pipeline has successfully promoted two-stage detectors, it is less harmonious with one-stage methods which are also important due to their competitive accuracy and computational efficiency.
As verified in \cref{fig1}, compared with Faster RCNN\cite{FasterRCNN}, FCOS\cite{Fcos} has a comparable supervised performance, but achieves a relatively limited improvement under the basic semi-supervised pipeline.
To figure out this problem, we analyze the core components of SSOD, e.g., pseudo-label selection and assignment.

\begin{figure*}[t]
    \centering
    \setlength{\abovecaptionskip}{5pt}
    \setlength{\belowcaptionskip}{-10pt}
    \includegraphics[width=\linewidth]{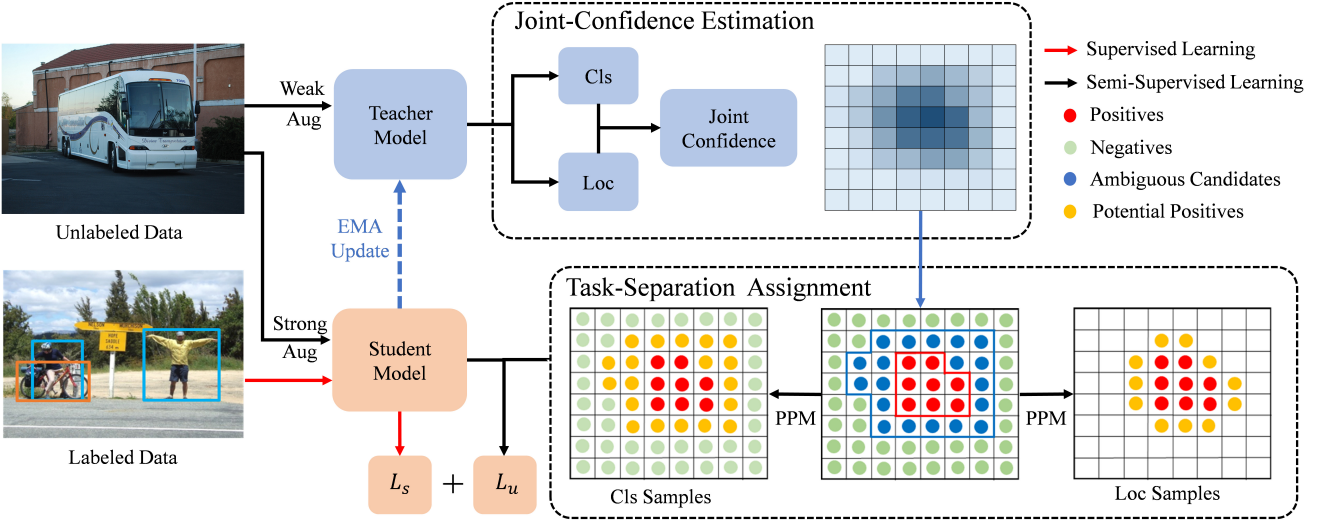}
    \caption{An overview of our Ambiguity-Resistant Semi-supervised Learning. Training batch contains both labeled and unlabeled images. On unlabeled images, the teacher first predicts the joint confidence via JCE. Then, TSA assigns and generates the training targets for the student. PPM denotes the potential positive mining in TSA. The overall loss consists of supervised $L_{s}$ and unsupervised loss $L_{u}$.}
    \label{fig2}
\end{figure*}

With comprehensive investigations in \cref{sec3.2}, we find that there exist selection and assignment ambiguities, hindering the semi-supervised learning of one-stage detectors.
The selection ambiguity denotes that the selected pseudo labels for unlabeled images are less accurate.
It is caused by the mismatch between classification scores and localization quality.
Specifically, compared with Faster RCNN, FCOS has a much smaller Pearson correlation coefficient between classification and localization (0.279 vs. 0.439), which is adverse to the pseudo-label selection.
The reason is that one-stage detectors like FCOS lack RPN\cite{FasterRCNN} and RoI Pooling/Align\cite{FasterRCNN, MaskRCNN} to extract accurate object information for localization quality estimation, meanwhile the predicted centerness of FCOS cannot properly represent the localization quality.

On the other hand, the assignment ambiguity indicates that samples on unlabeled images are assigned with improper labels.
Our experiments show that 73.5\% of positives are wrongly matched with negative labels, and there also exist many false positives.
In essence, the assignment strategy converts bounding boxes into pixel-level labels, but neglects the situations that many pseudo boxes are inaccurate and plenty of objects are missed due to the threshold filtering.
It causes the assignment ambiguity which misguides the detector.
Compared with two-stage detectors, one-stage detectors which require pixel-level labels, are more sensitive to the assignment ambiguity.

Based on these observations and analysis, we propose the \textbf{Ambiguity-Resistant Semi-supervised Learning} (ARSL) for one-stage detectors.
To mitigate the selection ambiguity, \textbf{Joint-Confidence Estimation} (JCE) is proposed to select high-quality pseudo labels based on the joint quality of classification and localization.
Specifically, JCE employs a double-branch structure to estimate the confidence of the two tasks, then combines them to format the joint confidence of detection results.
In training, the two branches are trained together in united supervision to avoid the sub-optimal state.
Different from other task-consistent or IoU-estimation methods\cite{Tood,GFL,IoUNet}, JCE explicitly integrates the classification and localization quality, and does not need complicated structures and elaborate learning strategies.
Additionally, JCE is more capable of picking high-quality pseudo labels and achieves a better SSOD performance, as verified in \cref{sec4.4}.


As for the assignment ambiguity, \textbf{Task-Separation Assignment} (TSA) is proposed to assign labels based on pixel-level predictions rather than unreliable pseudo boxes.
Concretely, based on the predicted joint confidence, TSA partitions samples into negatives, positives, and ambiguous candidates via the statistics-based thresholds.
The confident positives are trained on both classification and localization tasks, since they are relatively accurate and reliable.
While for the ambiguous candidates, TSA employs a ’divide-and-conquer’ strategy and separately exploits potential positives from them for the classification and localization task.
Compared with other dense-guided assignments\cite{DenseTeacher,DenseGuided,HumbleTeacher}, TSA adopts a more rational assignment metric and separately exploits positives for the two tasks, which can effectively mitigate the assignment ambiguity as proved in \cref{sec4.4}.
The general structure of ARSL is illustrated in \cref{fig2}, and our contributions are summarized as follows:
\begin{itemize}
\setlength{\itemsep}{0pt}
\setlength{\parskip}{0pt}
\item Comprehensive experiments are conducted to analyze the semi-supervised learning of one-stage detectors, and reveal that the limitation lies in the selection and assignment ambiguities of pseudo labels.
\item JCE is proposed to mitigate the selection ambiguity by jointly quantifying the classification and localization quality. To alleviate the assignment ambiguity, TSA separately exploits positives for the classification and localization task based on pixel-level predictions.
\item ARSL exhibits remarkable improvement over the basic SSOD baseline for one-stage detectors as shown in \cref{fig1}, and achieves state-of-the-art performance on MS COCO and PASCAL VOC.
\end{itemize}

\section{Related Works}
\label{sec2}

\noindent\textbf{Semi-Supervised Image Classification.} Semi-supervised classification has two dominant approaches: consistency regularization \cite{TemporalEns,MeanTea,UDA} and self-training\cite{Mixmatch,Fixmatch,PseudoLabel,NoisyStudent} (also known as pseudo-labeling). 
Consistency regularization forces the predictions to be invariant under various perturbations, e.g., different augmented inputs\cite{UDA}, ensemble predictions and models\cite{MeanTea}.
While in self-training, a pre-trained model is employed to predict pseudo labels for unlabeled data iteratively, and the model is optimized on both human-annotated and pseudo labels. 
NoisyStudent\cite{NoisyStudent} bolsters the robustness of student models by introducing proper noise into unlabeled data with pseudo labels.
FixMatch\cite{Fixmatch} further simplifies the self-training framework, in which one-hot pseudo labels are produced in weakly-augmented images and guide predictions on strongly-augmented views.
These effective technologies in image classification establish excellent foundations for semi-supervised detection.

\noindent\textbf{Semi-Supervised Object Detection.} In SSOD, the self-training and consistency based methods are inherited from semi-supervised image classification.
Following NoisyStudent\cite{NoisyStudent}, STAC\cite{stac} proposes a basic multi-stage pipeline, which first adopts a static teacher to generate labels for all unlabeled data and then trains the student.
To simplify the multi-stage process and produce high-quality labels, an end-to-end scheme\cite{UnbiasedTeacher,InstantTea,HumbleTeacher,SoftTea} is proposed to gradually update the teacher via the EMA of the student and predict pseudo labels online.
Under this scheme, many advanced studies further develop extensive approaches based on two-stage detectors.
Unbiased Teacher\cite{UnbiasedTeacher} tackles the pseudo-labeling bias via Focal Loss\cite{Focalloss}.
In Instant-Teaching\cite{InstantTea}, the student and teacher mutually rectify false predictions to alleviate the confirmation bias.
Humble Teacher\cite{HumbleTeacher} uses soft pseudo-labels for semi-supervised learning, which allows the student to distill richer information from the teacher.
Soft Teacher\cite{SoftTea} proposes the score-weighted classification loss and box jittering approach to spotlight reliable pseudo labels.
In this work, we focus on ameliorating the selection and assignment ambiguities in the semi-supervised learning of one-stage detectors.

\noindent\textbf{Selection Ambiguity.}
Selection ambiguity is caused by the inconsistency between the classification scores and localization quality.
Several existing methods\cite{RethinkingPseudoLabel,SoftTea,UnbiasedV2} in SSOD attempt to estimate the localization quality of pseudo labels via the uncertainty of bounding boxes.
In Rethinking Pseudo Labels\cite{RethinkingPseudoLabel}, box prediction is formulated as a classification task and localization quality is represented as the mean confidence of four boundaries.
Soft Teacher\cite{SoftTea} jitters the proposal boxes for several times (e.g. 10 times) and calculates the boundary variance as localization reliability. 
Unbiased Teacher V2\cite{UnbiasedV2} constructs a log-likelihood loss for regression task and guides the additional branches to predict the uncertainty of each boundary.
Compared with the aforementioned methods, our proposed JCE manifests two differences.
First, unlike the separate estimation for localization quality, JCE formulates a united representation of classification and localization, which avoids the sub-optimal state caused by separate training and is proved to be imperative in our ablation experiment.
Second, JCE maintains simplicity and flexibility, and is compatible with other prime tricks for localization, e.g., IoU-based losses.

\noindent\textbf{Assignment Ambiguity.}
For unlabeled data, inaccurate pseudo boxes and undetected objects match improper labels to samples, causing the assignment ambiguity.
Several methods attempt to alleviate the ambiguity of inaccurate pseudo boxes by selecting high-quality samples and pseudo boxes.
For instance, PseCo\cite{Pseco} chooses top-N performance samples as positives for each pseudo label.
LabelMatch\cite{LabelMatch} selects reliable pseudo labels via the matching degrees with adjacent results in NMS.
While for undetected objects, an efficacious idea is to directly transfer dense predictions of the teacher as pixel-level targets for consistency learning, including Humble Teacher\cite{HumbleTeacher}, Dense Teacher\cite{DenseTeacher}, and Dense Teacher Guidance\cite{DenseGuided}.
Compared with the aforementioned works, TSA integrates their advantages and further exploits potential positives for the classification and localization task, which is more robust to the assignment ambiguity.

\section{Methods}
\label{sec3}

To guarantee the generality, we take the classic FCOS\cite{Fcos} as an example to study the semi-supervised learning of one-stage detectors.
In \cref{sec3.1}, the basic SSOD framework is first applied to FCOS as our baseline.
Under this framework, the selection and assignment ambiguities of pseudo labels are analyzed in \cref{sec3.2}.
To mitigate the ambiguities, the proposed Joint-Confidence Estimation (JCE) is described in \cref{sec3.3}, and Task-Separation Assignment (TSA) is detailed in \cref{sec3.4}.

\subsection{Pseudo-Labeling Preliminary}
\label{sec3.1}

The advanced SSOD pipeline which follows the pseudo-labeling framework\cite{UnbiasedTeacher}, can be directly integrated into FCOS.
It consists of two stages: the burn-in stage and the self-training stage.
During the short burn-in stage, FCOS is pre-trained on the labeled data and duplicated into a student and teacher model.
In each iteration of the self-training stage, the teacher generates pseudo labels for unlabeled images and guides the student.
Specifically, the pseudo labels are predicted in the weakly-augmented views, and filtered according to their confidence which are obtained by multiplying the classification and centerness scores.
The retained pseudo labels are converted into pixel-level targets via the assignment strategy.
Then, the student is trained on labeled images and strongly-augmented unlabeled images with corresponding targets.
The overall loss $L$ of is formulated as a weighted sum of supervised loss $L_{sup}$ and unsupervised loss $L_{unsup}$:
\begin{equation}
    \label{equ1}
    L = L_{sup} + \beta L_{unsup},
\end{equation}
where $\beta$ indicates the unsupervised loss weight.
Finally, the teacher is updated based on the EMA of the student.

Nevertheless, FCOS obtains limited promotions under this pipeline. 
Compared with Faster RCNN which is a basic two-stage detector, there exists an improvement gap of approximately 2\% AP, as verified in \cref{fig1}.
With comprehensive investigations, we find that there exist ambiguities in pseudo-label selection and assignment, hindering the semi-supervised performance.
The detailed analysis is given in the following section.

\begin{table}[t]
    \caption{Comparison on pseudo labels predicted by Faster RCNN and FCOS. 'vanilla FCOS' denotes the FCOS without the centerness branch. 'Top-5 IoU' represents the mean IoU of top-5 detection results based on classification scores in each image. 'PCC' represents the Pearson Correlation Coefficient between the normalized classification scores and localization quality.}
    \centering
    \resizebox{0.95\linewidth}{!}{
    \begin{tabular}{lcccc}
    \hline
    Method & AP &  Mean IoU & Top-5 IoU & PCC  \\
    \hline
    Faster RCNN  & 26.4 & 0.348 & 0.641 & 0.439  \\
    vanilla FCOS & 25.2 & 0.369 & 0.585 & 0.235  \\
    FCOS  & 26.0 & 0.369 & 0.593 & 0.279  \\
    \hline
    \end{tabular}}
    \label{table1}
\end{table}

\subsection{Ambiguity Investigation}
\label{sec3.2}

In this part, we mainly investigate the quality of pseudo labels and assignment results in semi-supervised learning.
All detectors are trained on a standard 10\% split of COCO \emph{train2017} with a ResNet-50\cite{ResNet} backbone, and the statistics are obtained on COCO \emph{val2017}.

\noindent\textbf{Selection Ambiguity.}
The quality of pseudo labels in two-stage and one-stage detectors is investigated in \cref{table1}.
Since most one-stage detectors do not employ the centerness to re-calibrate classification scores, we first compare Faster RCNN with FCOS w/o Centerness in the second and third rows.
The mean IoU of detection results is 0.348 and 0.369 in Faster RCNN and FCOS, which indicates that FCOS has a slightly better localization ability.
Nevertheless, FCOS still performs worse on top-5 IoU selected by classification scores (0.585 vs. 0.641).
It demonstrates the weaker ability of FCOS to select high-quality pseudo labels.
Meanwhile, for the correlation between classification scores and localization quality, FCOS has a much smaller PCC than Faster RCNN (0.235 vs. 0.439).
On the other hand, as shown in the fourth row, the auxiliary centerness brings limited improvement on top-5 IoU (0.585 vs. 0.593) and PCC (0.235 vs. 0.279), and there still exists a large gap with Faster RCNN.
These statistics reveal that there exists a more serious inconsistency between classification and localization in FCOS.
Consequently, this mismatch affects the selection of high-quality pseudo labels, suppressing the semi-supervised performance.

\begin{figure}[t]
    \centering
    \setlength{\belowcaptionskip}{-10pt}
    \includegraphics[width=\linewidth]{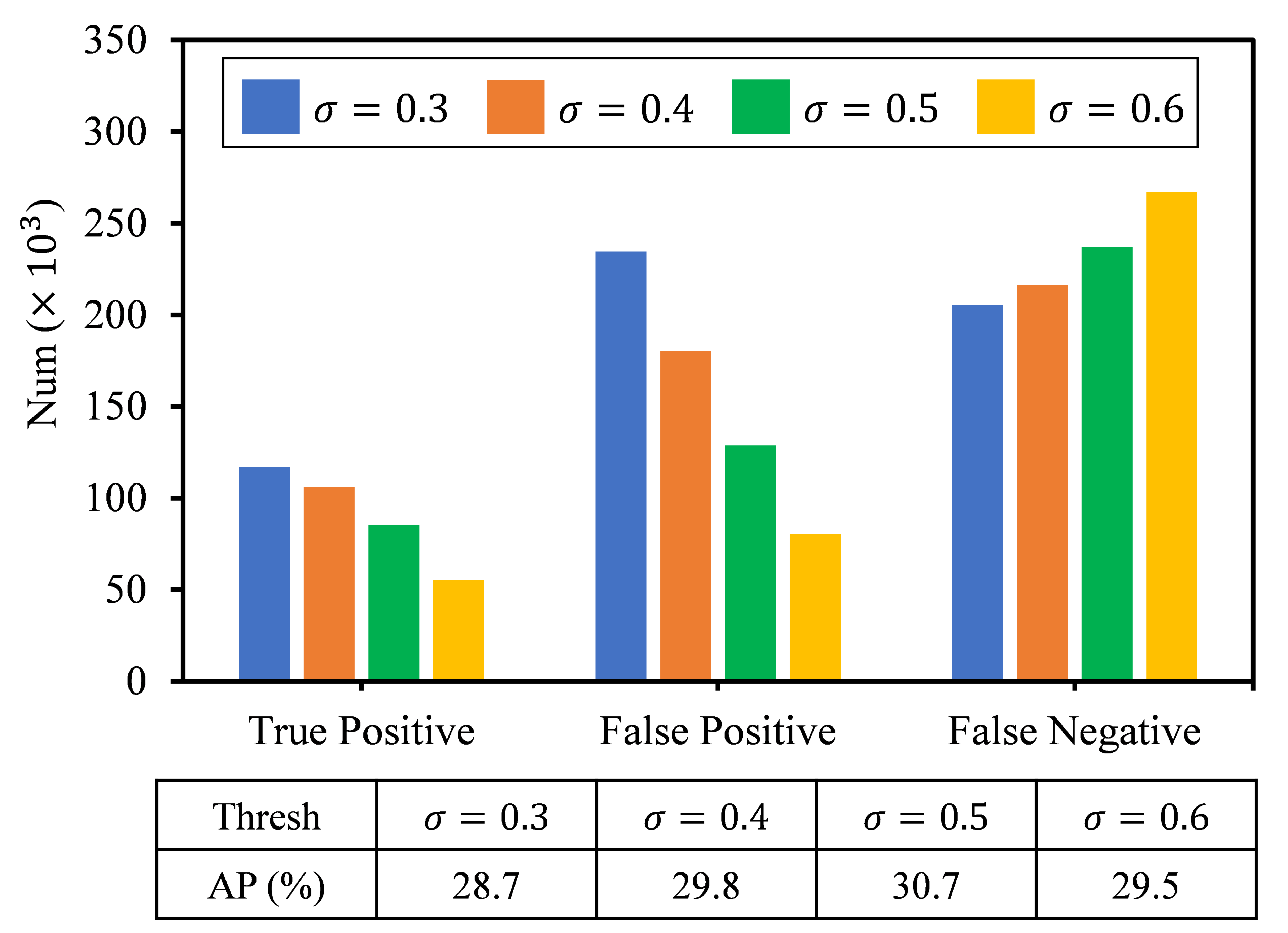}
    \caption{Investigation on the assignment ambiguity of FCOS under different filtering thresholds $\sigma$. The assignment results are obtained based on selected pseudo labels.}
    \label{fig3}
\end{figure}

\noindent\textbf{Assignment Ambiguity.}
To analyze the rationality of label assignment, we count assignment ambiguities of FCOS under different filtering thresholds.
As shown in \cref{fig3}, a low filtering threshold retains more pseudo boxes and covers more true-positive samples, while a high threshold avoids more false positives.
When setting the threshold to 0.5, the detector achieves a relatively proper trade-off between true and false positives, obtaining the best semi-supervised performance.
Nevertheless, under this condition, about 73.5\% of positives are incorrectly assigned as negatives (false negatives).
Meanwhile, there also exist a large amount of false positives which confuse the detector.
Substantially, the quality of assignment results depends on reliable bounding boxes.
However, many pseudo boxes are inaccurate and plenty of objects are missed due to the threshold filtering.
The box-based assignment is not robust to these situations, which causes the assignment ambiguity.

\subsection{Joint-Confidence Estimation}
\label{sec3.3}



It is observed in \cref{sec3.2} that the inconsistency between classification and localization causes the selection ambiguity of pseudo labels.
To this end, we propose a simple and effective method named Joint-Confidence Estimation (JCE) to resist the selection ambiguity.

\begin{figure}[t]
    \centering
    \setlength{\abovecaptionskip}{5pt}
    \setlength{\belowcaptionskip}{-10pt}
    \includegraphics[width=\linewidth]{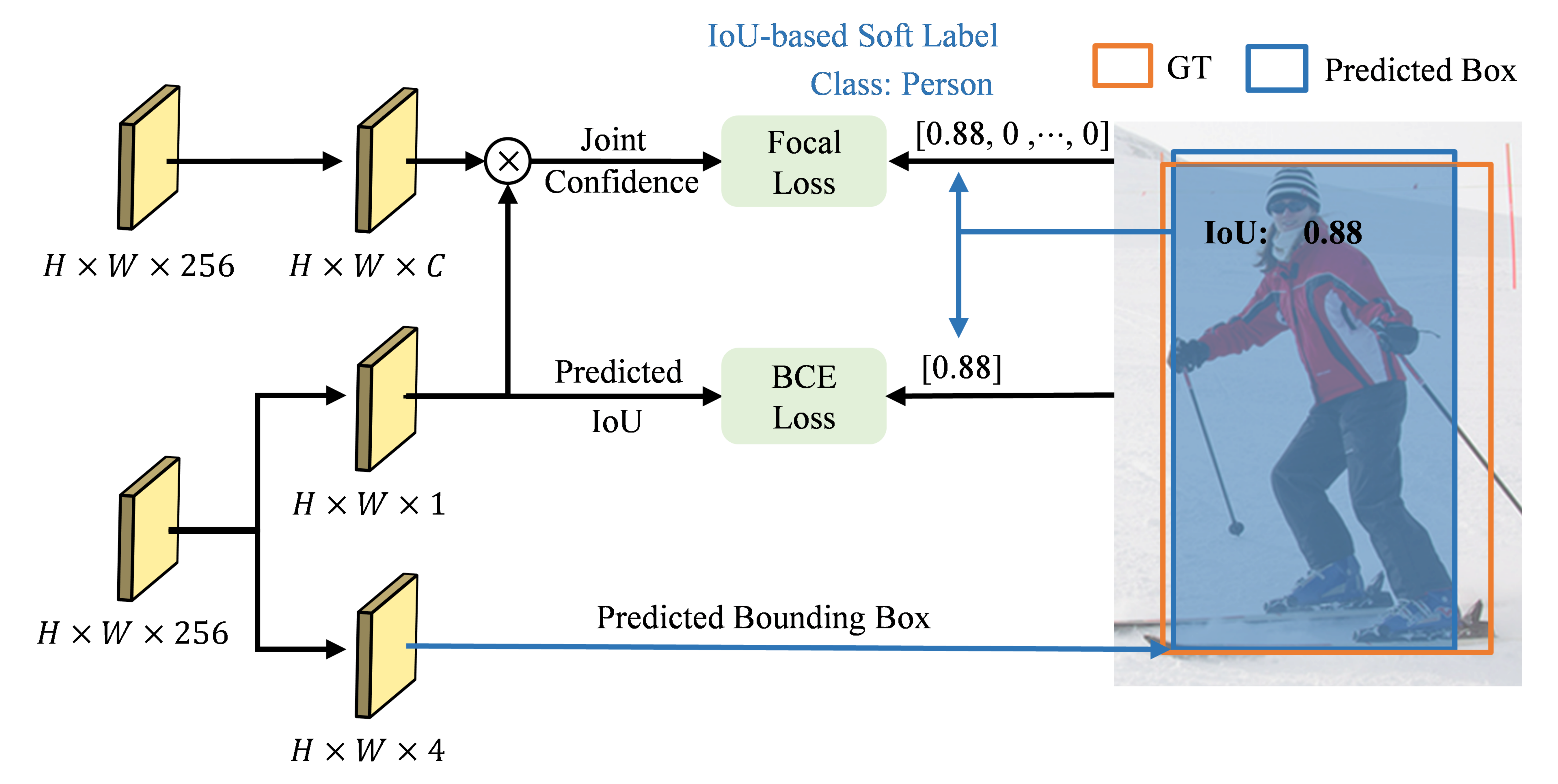}
    \caption{Joint-confidence Estimation. Two branches of JCE are trained together, and the IoU between predicted boxes and corresponding GT are employed to generate IoU-based soft label.}
    \label{fig4}
\end{figure}

The gist of JCE is to format a joint confidence of the classification and localization for pseudo-label selection.
To achieve this, JCE employs a double-branch structure, including the original classification branch to recognize object categories and the auxiliary branch to estimate the localization quality, as shown in \cref{fig4}.
The joint confidence $\hat{S}$ is obtained by combining the classification scores $\hat{S}_{cls}$ and the predicted IoU $\hat{S}_{iou}$, as:
\begin{equation}
\label{equ2}
	\hat{S} = \hat{S}_{cls} * \hat{S}_{iou}.
\end{equation}
To avoid the sub-optimal state caused by separate training, the two tasks are merged to format the united supervision.
The united classification objective $L_{cls}$ is calculated based on Focal Loss:
\begin{equation}
\label{equ3}
    L_{cls} = \text{FL}(\hat{S},S).
\end{equation}
For labeled images, the learning targets $S$ are the IoU-based soft label proposed in VFNet\cite{VFL} and GFL\cite{GFL}.
While for unlabeled images, since the pseudo labels are unreliable, learning the IoU between student's predictions and pseudo boxes is less sensible.
Therefore, the learning targets $S$ are set according to the teacher's predictions and extended to:
\begin{equation}
\label{equ4}
	{S} = 
	\begin{cases}
	\{0, \cdots, IoU, \cdots, 0\},              & \text{Labeled} \\
	\{0, \cdots, Max(\hat{S}_{t}), \cdots, 0\}, & \text{Unlabeled} \\
	\end{cases}
\end{equation}
where the soft label is on the corresponding class channel, $IoU$ represents the IoU between predicted boxes and corresponding GT boxes, and $Max(\hat{S}_{t})$ is the largest score of the teacher's responses among all categories.

Moreover, to make the auxiliary branch focus on IoU estimation, an additional IoU loss $L_{iou}$ is added as:
\begin{equation}
\label{equ5}
	L_{iou} = BCE(\hat{S}_{iou},IoU),
\end{equation}
where $BCE$ denotes the binary cross entropy loss.

As verified in \cref{sec4.4}, the proposed joint confidence effectively mitigates the selection ambiguity and bolsters the semi-supervised performance.
Note that JCE can be directly applied to the FCOS baseline without changing the network structure.
For other one-stage detectors, it only adds a lightweight $3\times3$ convolution layer, which maintains simplicity and efficiency.

\begin{figure}[t]
    \centering
    \setlength{\abovecaptionskip}{0pt}
    \setlength{\belowcaptionskip}{-10pt}
    \includegraphics[width=\linewidth]{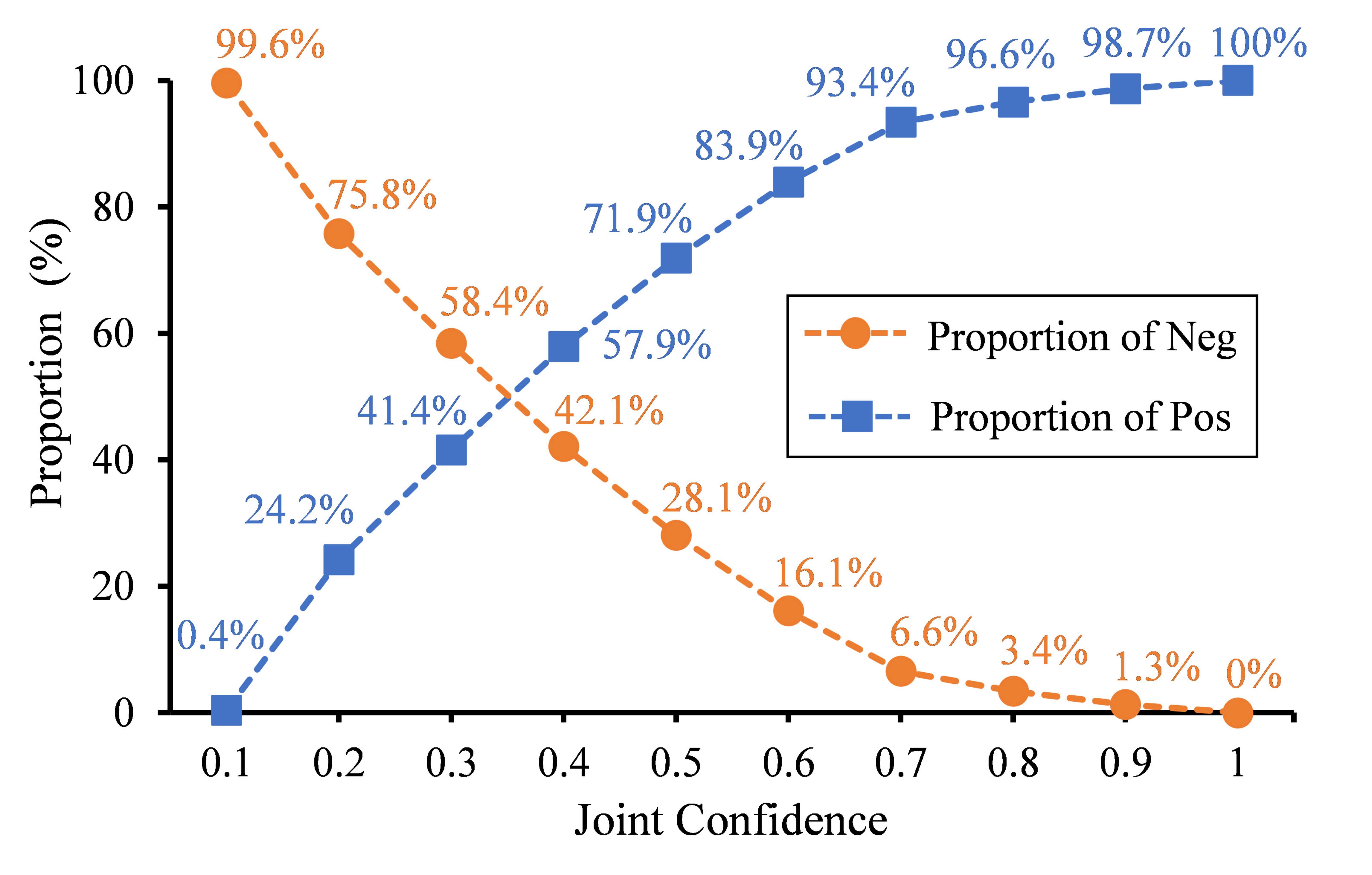}
    \caption{Proportion between positives and negatives in Joint-confidence intervals.}
    \label{fig5}
\end{figure}

\subsection{Task-Separation Assignment}
\label{sec3.4}

As proved in \cref{sec3.2}, conducting label assignment based on pseudo boxes provokes the assignment ambiguity.
To tackle this problem, we intend to assign labels based on the proposed joint confidence rather than unreliable pseudo boxes, since the joint confidence predicted by the teacher can quantify the quality of samples.
However, as shown in \cref{fig5}, though samples with high and low confidence are highly likely to be positives and negatives respectively, the samples in the middle regions are still ambiguous.
To this end, Task-Separation Assignment (TSA) is proposed to ease the assignment ambiguity.
TSA employs a 'divide-and-conquer' strategy, and separately exploits potential positives from ambiguous samples for the classification and localization, since the two tasks have different sensitivity to the ambiguity.


Specifically, TSA uses negative and positive thresholds $\{\tau_{neg}, \tau_{pos}\}$ to divide samples into negatives, ambiguous candidates, and positives, as follows:
\begin{equation}
\label{equ6}
	{x_i} = 
	\begin{cases}
	\text{Negative},  &  Max(\hat{S}_{i}) < \tau_{neg} \\
	\text{Candidate}, & \tau_{neg} \leq  Max(\hat{S}_{i}) \leq \tau_{pos} \\
	\text{Positive},  &  Max(\hat{S}_{i}) > \tau_{pos} \\
	\end{cases},
\end{equation}
where $\hat{S}_i$ is the joint confidence predicted by the teacher of $i$-th sample.
$\tau_{neg}$ is fixed to 0.1, and $\tau_{pos}$ is dynamically calculated based on the mean and standard deviation of candidates and positives:
\begin{equation}
\label{equ7}
	\tau_{pos} = {Max(\hat{S})}_{mean} + {Max(\hat{S})}_{std}.
\end{equation}
The confident positives are trained on both classification and localization tasks, since they are relatively accurate and reliable.
TSA further exploits potential positives from ambiguous samples for the two tasks, respectively.

\begin{table*}[t]
    \setlength{\belowcaptionskip}{-20pt}
    \caption{Experimental results on COCO-Standard. Two-stage detectors employ Faster RCNN as the baseline, while FCOS is used for one-stage detectors. ${}^{*}$ and ${}^{\dag}$ denotes the additional patch-shuffle and large scale jittering augmentation respectively.}
    \centering
    \resizebox{0.85\linewidth}{!}{
    \begin{tabular}{lccccc}
    \hline
    \multirow{2}{*}{Methods} & \multirow{2}{*}{Reference} & \multicolumn{4}{c}{COCO-Standard} \\
    \cline{3-6}
    & & 1\% & 2\% & 5\% & 10\% \\
    \hline
    Faster RCNN\cite{FasterRCNN} (Supervised) & - &  $10.02\pm0.38$  &  $15.04\pm0.31$  &  $20.82\pm0.13$  &  $26.44\pm0.11$ \\
    STAC\cite{stac}  &  arXiv20  & $13.97\pm0.35$  &  $18.25\pm0.25$  &  $24.38\pm0.12$  &  $28.64\pm0.21$ \\
    ISMT\cite{ISMT}  & CVPR21 & $18.88\pm0.74$  &  $22.43\pm0.56$  &  $26.37\pm0.24$  &  $30.53\pm0.52$ \\
    Humble Teacher\cite{HumbleTeacher}  & CVPR21 &  $16.96\pm0.38$  &  $21.72\pm0.24$  &  $27.70\pm0.15$  &  $31.61\pm0.28$ \\
    Unbiased Teacher\cite{UnbiasedTeacher}  & ICLR21 &  $20.75\pm0.12$  &  $24.30\pm0.07$  &  $28.27\pm0.11$  &  $31.50\pm0.10$ \\
    Active Teacher\cite{ActiveTea} & CVPR22 & 22.20 & 24.99 & 30.07 & 32.58 \\
    Unbiased Teacher V2\cite{UnbiasedV2}  & CVPR22 &  $21.84\pm0.13$  &  $26.14\pm0.01$  &  $30.06\pm0.14$  &  $33.50\pm0.03$ \\
    $\text{Soft Teacher}^{\dag}$\cite{SoftTea}  & ICCV21 &  $20.46\pm0.39$  &  -  &  $30.74\pm0.08$  &  $34.04\pm0.14$ \\
    PseCo\cite{Pseco}  & ECCV22 &  $22.43\pm0.36$  &  $27.77\pm0.18$  &  $32.50\pm0.08$  &  $36.06\pm0.24$ \\
    \hline
    FCOS\cite{Fcos} (Supervised) & - & $9.05\pm0.31$ & $14.40\pm0.28$ & $20.69\pm0.22$ & $26.01\pm0.15$ \\
    Unbiased Teacher V2\cite{UnbiasedV2}  & CVPR22 &  $22.71\pm0.42$  &  $26.03\pm0.12$  &  $30.08\pm0.04$  &  $32.61\pm0.03$ \\
    Dense Teacher\cite{DenseTeacher}  & ECCV22 &  $19.64\pm0.34$  &  $25.39\pm0.13$  &  $30.83\pm0.21$  &  $35.11\pm0.13$ \\
    $\text{DSL}^{*}$\cite{Dsl}  & CVPR22 &  $22.03\pm0.28$  &  $25.19\pm0.37$  &  $30.87\pm0.24$  &  $36.22\pm0.18$ \\
    \textbf{ARSL (FCOS)} & - &  $22.82\pm0.26$  &  $28.11\pm0.19$  &  $33.14\pm0.12$  &  $36.90\pm0.03$ \\ 
    $\textbf{ARSL}^{\dag}$ \textbf{(FCOS)}  & - &  $25.36\pm0.32$   &  $29.08\pm0.21$  &  $34.45\pm0.16$  &   $38.50\pm0.05$ \\ 
    $\textbf{ARSL}^{\dag}$ \textbf{(RetinaNet)}  & - &  $25.16\pm0.25$   &  $28.68\pm0.24$  &  $34.30\pm0.21$  &   $38.42\pm0.03$ \\     
    \hline
    \end{tabular}}
    \label{table2}
\end{table*}

\noindent\textbf{Classification Mining.}
The candidate samples are composed of low-confidence positives and hard negatives.
For the classification task, though these candidates usually involve background regions, they are not easy background as verified in \cref{sec4.3}, and also contain partial foreground information which is worth learning.
Therefore, all the candidate samples participate in the consistency learning to mimic the classification responses of the teacher.


\noindent\textbf{Localization Mining.}
The localization task is more rigorous and sensitive in sample selection, since excessive discrepancy among samples disturbs the optimization of the locator.
With this consideration, we select potential positives according to their similarity with positives, and set the matching positives as localization targets.
The similarity metric contains several factors: 
(1) Classification similarity. Candidate samples should have the same predicted category with positives. 
(2) Localization similarity. The IoU between candidate boxes and positive boxes should be larger than the threshold (0.6 by default). 
(3) Position similarity. The location of candidate samples should be inside the positive boxes.
The candidates which successfully match positive samples, are selected as potential positives in the localization task.
Given a potential positive sample, its localization target $B$ is calculated based on the weighted average of matched positives:
\begin{equation}
\label{equ8}
B = \frac{\sum_{i=1}^N Max(\hat{S_i})*\hat{B_i}} {\sum_{i=1}^N Max(\hat{S_i})},
\end{equation}
where $N$ represents the number of matched positives, $\hat{S_i}$ and $\hat{B_i}$ are the joint confidence and bounding box of $i$-th positives predicted by the teacher.

\noindent\textbf{Loss Function.}
The overall unsupervised loss $L_{unsup}$ consists of three parts:
\begin{align}
\label{equ9}
    L_{unsup} &= \frac{1}{N_{cls}}\sum_{i=1}^{N_{cls}} L_{cls}(s_i,\hat{s}_i)
    + \frac{1}{N_{loc}}\sum_{i=1}^{N_{loc}} L_{loc}(b_i,\hat{b}_i) \nonumber \\ 
    &+ \frac{\lambda}{N_{loc}}\sum_{i=1}^{N_{loc}} L_{iou}(p_i,\hat{p}_i),
\end{align}
where $N_{cls}$ and $N_{loc}$ are the number of samples for classification and localization, $L_{cls}$ and $L_{iou}$ are defined in \cref{equ3} and \cref{equ5}, $L_{loc}$ denotes the GIoU loss \cite{GIoU}, and the weighting terms $\lambda$ is set to 0.5.



\section{Experiments}
\label{sec4}

\subsection{Experiments and Implementation Details}
\label{sec4.1}

\noindent\textbf{Experiment Settings.}
The experiments are conducted on the MS COCO\cite{Mscoco} benchmark and PASCAL VOC\cite{Pascalvoc} datasets.
MS COCO contains 80 classes with 118$k$ labeled images and 123$k$ unlabeled images.
VOC2007 has 5$k$ training images from 20 classes and another 5$k$ images for testing, while VOC2012 has 11$k$ labeled images.
Following previous works, the proposed method is examined on three experimental scenarios: 
(1) \textbf{COCO-Standard}. 1\%, 2\%, 5\%, and 10\% of the \emph{train2017} set are randomly sampled as labeled data, and the remaining images are regarded as unlabeled data.
For each split, we create 5 data folds and report the mean $AP_{50:90}$ on the \emph{val2017}.
(2) \textbf{COCO-Full}. COCO-Full utilizes the \emph{train2017} as labeled data and \emph{unlabel2017} as unlabeled data. 
The COCO standard $AP_{50:90}$ is adopted as the evaluation metric.
(3) \textbf{VOC}. As for VOC, the trainval sets of VOC2007 and VOC2012 are employed as labeled data and unlabeled data. respectively. The models are validated in the VOC2007 test set, and the $AP_{50:90}$ along with $AP_{50}$ are reported as the evaluation metrics.

\noindent\textbf{Implementation Details.}
We adopt the widely used FCOS\cite{Fcos} as the baseline and ResNet-50\cite{ResNet} pretrained on ImageNet\cite{Imagenet} as the backbone.
All the models are trained on 8 GPUs with 8 images per GPU (4 labeled and 4 unlabeled images) and optimized with SGD.
Weight decay and momentum are set to 0.0001 and 0.9, respectively.
The base learning rate is set to 0.02 without the decay scheme in all our experiments.
For COCO-standard and VOC experiments, the models are trained for 90K iterations, and the learning schedule is extended to 360K iterations for the COCO-full setting.
For a fair comparison, following previous works, weak augmentation only contains random flip, while strong augmentation includes random flip, color jittering, and cutout unless specified.
The 'burn-in' strategy is also applied to initialize the model before semi-supervised learning.
The weight of unsupervised loss is set to 2.
The teacher model is updated through EMA with a momentum of 0.9996.

\begin{table}[t]
    \setlength{\belowcaptionskip}{-10pt}
    \caption{Experimental results on COCO-Full. Note that $1\times$ indicates $90k$ training iterations, and $N\times$ is $N\times90k$ iterations.}
    \centering
    \resizebox{\linewidth}{!}{
    \begin{tabular}{lc}
    \hline
    Methods & COCO-Full (100\%) \\
    \hline
    STAC\cite{stac} ($6\times$) & $39.48\xrightarrow{-0.27}39.21$ \\
    Unbiased Teacher\cite{UnbiasedTeacher} ($3\times$) & $40.20\xrightarrow{+1.10}41.30$ \\
    Soft Teacher\cite{SoftTea} ($8\times$) & $40.90\xrightarrow{+3.60}44.50$ \\
    Unbiased Teacher V2\cite{UnbiasedV2} ($8\times$) & $40.90\xrightarrow{+3.85}44.75$ \\
    PseCo\cite{Pseco} ($8\times$) & $41.00\xrightarrow{+5.10}46.10$ \\
    \hline
    Dense Teacher\cite{DenseTeacher} ($6\times$) & $41.22\xrightarrow{+3.72}44.94$ \\
    DSL\cite{Dsl} ($4\times$) & $40.20\xrightarrow{+3.60}43.80$ \\
    \textbf{ARSL} ($4\times$) & $40.40\xrightarrow{+4.70}45.10$ \\
    \hline
    \end{tabular}}
    \label{table3}
\end{table}

\begin{table}[t]
    \caption{Experimental results on VOC protocol.}
    \centering
    \begin{tabular}{lcc}
    \hline
    Methods & $AP_{50}$ & $AP_{50:95}$ \\
    \hline
    Faster RCNN\cite{FasterRCNN} (Supervised) & 72.75 & 42.04 \\
    CSD\cite{CSD} & 74.7 & - \\
    STAC\cite{stac} & 77.45 & 44.64 \\
    ISMT\cite{ISMT} & 77.23 & 46.23 \\
    Unbiased Teacher\cite{UnbiasedTeacher} & 77.37 & 48.69 \\
    Instant Teaching\cite{InstantTea} & 79.20 & 50.00 \\
    \hline
    FCOS\cite{Fcos} (Supervised) & 71.36 & 45.52  \\
    Dense Teacher\cite{DenseTeacher} & 79.89 & 55.87 \\
    \textbf{ARSL} & 80.40 & 56.40 \\
    \hline
    \end{tabular}
    \label{table4}
\end{table}

\subsection{Comparison with State-of-the-Arts}
\label{sec4.2}

Under all three protocols, the proposed ARSL is compared with existing SOTA methods including both two-stage and one-stage detectors.

\noindent\textbf{COCO-Standard.}
Under the COCO-standard protocol, the results are given in \cref{table2}. 
The FCOS baseline achieves comparable performance with Faster RCNN, exhibiting the fairness of semi-supervised comparison.
For all splits, the proposed ARSL derives impressive improvements over the supervised baseline and outperforms all the two-stage and one-stage methods, which demonstrates its effectiveness.
When further adopting the large-scale jittering for augmentation, ARSL establishes the new state-of-the-art performance.
Moreover, ARSL also achieves remarkable SSOD performance on RetinaNet\cite{Focalloss}, which verifies its generality on both anchor-based and anchor-free one-stage detectors.

\noindent\textbf{VOC \& COCO-Full.}
The results on the COCO-full setting are shown in \cref{table3}.
Since the baseline reported in previous works are different and the learning schedules vary a lot, we report SSOD performance along with the supervised baseline and mainly compare the improvements.
The proposed ARSL achieves a remarkable improvement (4.70\% AP) under a relatively short learning schedule, exhibiting its superiority.
As for VOC, the results are reported in \cref{table4}.
ARSL ameliorates the supervised baseline by 9.04\% and 10.88\% on $AP_{50}$ and $AP_{50:95}$, achieving competitive performance with existing works.

\begin{table}[t]
    \setlength{\belowcaptionskip}{-10pt}
    \caption{The impacts of components on detection performance. JCE, TSA indicate the proposed Joint-Confidence Estimation and Task-Separation Assignment.}
    \centering
    \begin{tabular}{lccc}
    \hline
    Methods & $AP$ & $AP_{50}$ & $AP_{75}$ \\
    \hline
    FCOS (Supervised)      & 26.0 & 43.6 & 26.7 \\
    FCOS (Semi-Supervised) & 30.7 & 47.1 & 32.4 \\
    \hline
    $+$ JCE              & 34.7 & 52.4 & 37.3 \\
    $+$ TSA (w/o mining) & 35.6 & 54.3 & 38.1 \\
    $+$ TSA (w/ mining)  & 36.9 & 55.4 & 39.6 \\
    \hline
    \end{tabular}
    \label{table5}
\end{table}

\begin{table}[t]
    \caption{Ablation studies on Joint-Confidence Learning. 'United Supervision' indicates the joint training of the IoU-prediction and classification task. 'Specific targets' denotes that the classification targets of unlabeled data is set as max responses of the teacher.}
    \centering
    \begin{tabular}{lr}
    \hline
    Strategies of JCE & $AP$ \\
    \hline
    baseline & $30.7$ \\
    $+$ IoU prediction & $32.0(+1.3)$ \\
    $+$ United supervision & $34.2(+2.2)$ \\
    $+$ Specific targets for unlabeled data & $34.7(+0.5)$ \\
    \hline
    \end{tabular}
    \label{table6}
\end{table}


\subsection{Ablation Studies}
\label{sec4.3}

To provide a better understanding of the proposed method, we first assess the influence of each component on detection performance, then analyze their details in the following.
All experiments are conducted on the 10\% split of COCO-standard.

\noindent\textbf{Component Impact.}
The effectiveness of components is reported in \cref{table5}.
FCOS under the basic SSOD framework described in \cref{sec3.1} obtains 30.7\% AP.
With JCE, the accuracy is boosted to 34.7\% AP, delivering a remarkable improvement of 4.0\% AP.
It demonstrates the superiority of JCE compared with the original centerness scheme.
When applying TSA and simply ignoring the ambiguous candidates, the performance is increased by 0.9\% AP (34.7\% AP vs. 35.6\% AP).
This substantiates that assigning labels based on dense predictions rather than pseudo boxes is more rational.
By further mining the positives from candidates, TSA further bolsters the performance to 36.9\% AP, which demonstrates the effectiveness of TSA.
Compared with the SSOD baseline, the proposed method achieves an overall improvement of 6.2\% AP (30.7\% AP vs. 36.9\% AP).

\noindent\textbf{Strategies on JCE.}
\cref{table6} shows the ablation studies on different strategies of JCE. 
The performance is increased from 30.7\% AP to 32.0\% AP by replacing the centerness estimation of FCOS with IoU prediction.
The united supervision which avoids the sub-optimal state caused by separate training, improves the performance to 34.2\% AP.
Such a large gain (2.2\% AP) demonstrates the effectiveness of the united training.
Setting specific targets for unlabeled data further ameliorates the performance to 34.7\% AP.


\begin{table}[t]
    \caption{Quality Analysis of Potential Positives. 'Mean IoU' represents the average IoU between samples and corresponding GTs. 'Percent' indicates the proportion of potential positives in candidate sample.}
    \centering
    \begin{tabular}{lcc}
    \hline
    Type & Mean IoU & Percent\\
    \hline
    Potential positives for cls & 0.369 & 100\% \\
    Potential positives for loc & 0.504 & 33.9\% \\
    Learning targets for loc    & 0.633 & -  \\
    \hline
    \end{tabular}
    \label{table7}
\end{table}

\noindent\textbf{Quality Analysis of TSA.}
\cref{table7} analyzes the quality of potential positives exploited by TSA.
For the classification task, all candidates are regarded as potential positives and have a mean IoU of 0.369, which verifies that they are not easy backgrounds and worth learning.
As for the localization task, 33.9\% of candidates are selected as potential positives with a mean IoU of 0.504, and their learning targets obtain a mean IoU of 0.633. 
It indicates that our matching strategy does select high-quality samples from candidates.

\begin{table}[t]
    \setlength{\belowcaptionskip}{-5pt}
    \caption{Selection Ambiguity Mitigation. 'T-Head' denotes the task-aligned head in TOOD and QFL is the quality focal loss in GFL. The metrics follow the settings presented in \cref{sec3.2}. The statistics are calculated by the final model of 10\% split on validation set.}
    \centering
    \begin{tabular}{lccc}
    \hline
    Methods & Top-5 IoU & PCC & AP\\
    \hline
    FCOS           & 0.614 & 0.299 & 30.7 \\
    FCOS w/ T-head\cite{Tood} & 0.632 & 0.361 & 31.9 \\
    FCOS w/ QFL\cite{GFL}    & 0.628 & 0.353 & 32.3 \\
    FCOS w/ JCE    & 0.656 & 0.395 & 34.7 \\ 
    \hline
    \end{tabular}
    \label{table8}
\end{table}

\subsection{Ambiguity Mitigation}
\label{sec4.4}

\noindent\textbf{Selection Ambiguity.} 
The influence of the proposed JCE on the ambiguity mitigation is verified in \cref{table8}.
The metrics follow the settings presented in \cref{sec3.2}.
Compared with the FCOS baseline, the Top-5 IoU is improved from 0.614 to 0.656, and PCC is increased by 0.096 (0.299 vs. 0.395), which substantiates that JCE can effectively mitigate the selection ambiguity.
Moreover, JCE is also compared with existing effective methods that have been proven to ease the prediction inconsistency in supervised learning.
For T-head and GFL, our JCE obtains a higher PCC and achieves a larger improvement in semi-supervised learning.

\begin{figure}[t]
  \centering
  \setlength{\abovecaptionskip}{5pt}
  \setlength{\belowcaptionskip}{-15pt}
   \includegraphics[width=\linewidth]{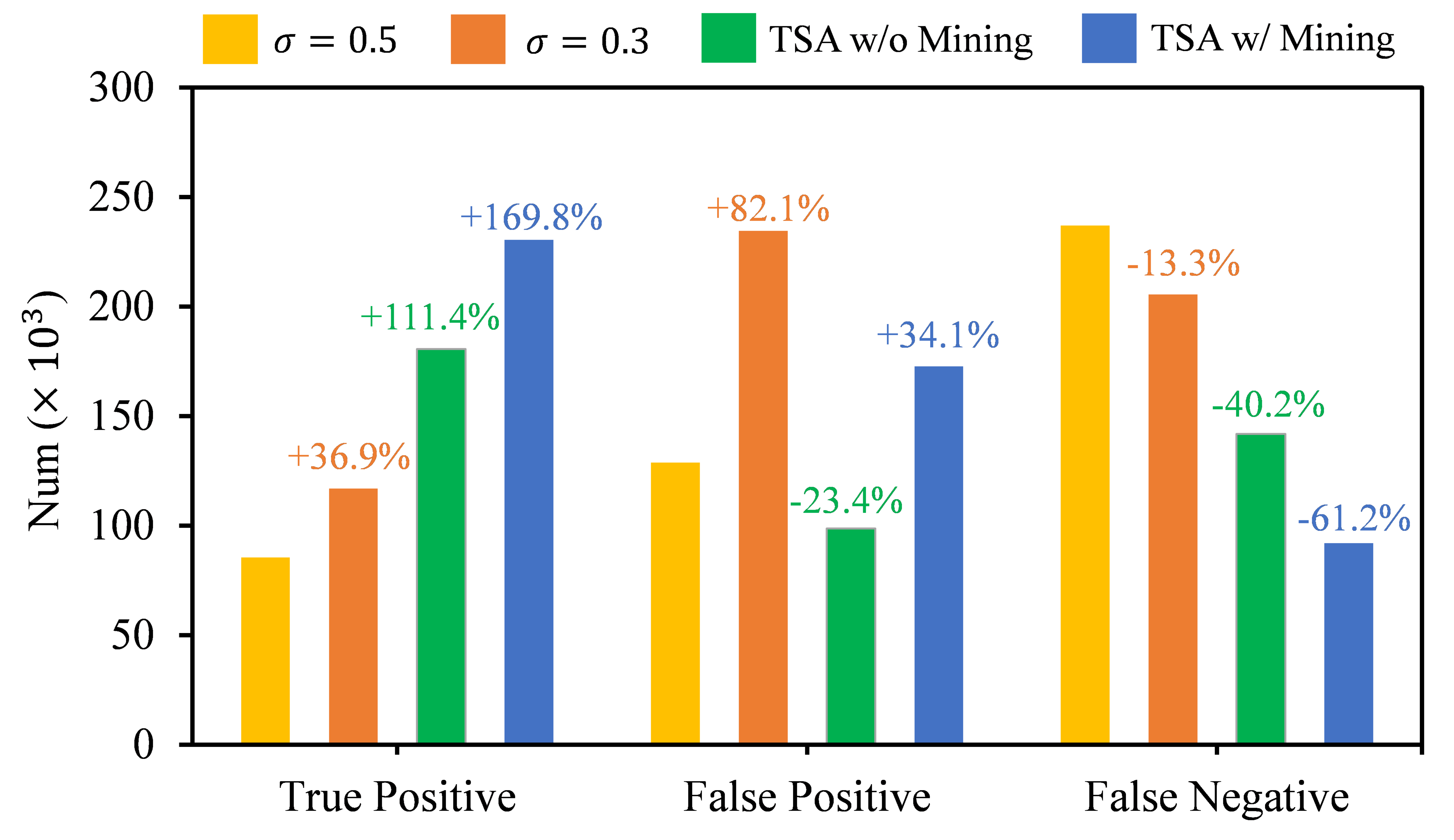}
   \caption{Mitigation of Assignment Ambiguity. $\sigma$ indicates the filtering threshold of pseudo boxes. The statistics are counted on the COCO validation set.}
   \label{fig6}
\end{figure}

\noindent\textbf{Assignment Ambiguity.}
We also analyze the effectiveness of TSA on assignment ambiguity, as shown in \cref{fig6}.
In the box-based assignment, though the decline of threshold increases the number of true positives ($+36.9\%$), false positives are also grown by 82.1\%.
While under the TSA without potential positive mining, true positives are significantly bolstered by 111.4\% and false positives are depressed by 23.4\%, which verifies that assignment based on the joint confidence are more robust to inaccurate pseudo boxes and missed objects in SSOD.
Exploiting potential positives further boosts the true positives by 58.4\%, obtaining an overall increase of 169.8\% and a total decrease on false negatives of 61.2\%.
It reflects that TSA does exploit many true positives from ambiguous candidates.
Note that the slight increase of false positives is caused by that all ambiguous candidates are regarded as positives in the classification task.
These observations reveal that the proposed TSA can mitigate the assignment ambiguity to a large extent.

\section{Conclusion}
\label{sec5}

In this study, we investigate the selection and assignment ambiguity in the semi-supervised learning of one-stage detectors.
To mitigate these ambiguities, the Ambiguity-Resistant Semi-supervised Learning (ARSL) is proposed, consisting of Joint-Confidence Estimation and Task-Separation Assignment.
The verification experiments demonstrate that our methods can effectively alleviate the ambiguities.
Compared with the baseline, ARSL obtains a remarkable improvement and achieves state-of-the-art performance on MS COCO and PASCAL VOC.

~

\noindent\textbf{Acknowledgments.}
This work was supported by the National Natural Science Foundation of China [grant numbers 61991415 and 62225308] and National Key R\&D Program of China [grant numbers 2020YFC1521703].

{\small
\bibliographystyle{ieee_fullname}
\bibliography{ARSL}
}

\end{document}